\documentclass[copyright,noncommercial]{eptcs}
\pdfoutput=1
\providecommand{\volume}{5}
\providecommand{\anno}{2009}
\providecommand{\firstpage}{27}
\providecommand{\eid}{3}
\providecommand{\event}{Y. Deville and C. Solnon (Eds): Sixth International Workshop\\
 on Local Search Techniques in Constraint Satisfaction (LSCS'09).}

\usepackage{algorithm}
\usepackage{amsmath}
\usepackage[british]{babel}
\usepackage{color}
\usepackage{graphicx}

\newcommand{\RF}{\mathit{RF}}
\newcommand{\RAF}{\mathit{RAF}}
\newcommand{\RWF}{\mathit{RWF}}
\newcommand{\Sr}{\mathit{S_r}}
\newcommand{\SFr}{\mathit{SF_r}}
\newcommand{\PFr}{\mathit{PF_r}}
\newcommand{\RC}{\mathit{RC}}

\newcommand{\capacity}{\mathit{cap}}
\newcommand{\demand}{\mathit{demand}}
\newcommand{\holding}{\mathit{deltaT}}
\newcommand{\maxDelay}{\mathit{g}}
\newcommand{\maxIter}{\mathit{maxIter}}
\newcommand{\now}{\mathit{now}}
\newcommand{\violations}{\mathit{violations}}

\newcommand{\card}[1]{\mathit{|#1|}}

\title{Dynamic Demand-Capacity Balancing \\
  for Air Traffic Management \\
  Using Constraint-Based Local Search: \\
  First Results
}

\author{
  Farshid Hassani Bijarbooneh \qquad\qquad
  Pierre Flener \qquad\qquad
  Justin Pearson
  \institute{Department of Information Technology \\
    Uppsala University,
    Box 337, SE -- 751 05 Uppsala, Sweden}
  \email{\{Pierre.Flener,Justin.Pearson\}@it.uu.se}
}

\def\titlerunning{Dynamic Demand-Capacity Balancing for ATM Using CBLS: First Results}
\def\authorrunning{F. Hassani Bijarbooneh, P. Flener, and J. Pearson}

\begin{document}

\maketitle

\begin{abstract}
  Using constraint-based local search, we effectively model and efficiently solve the problem of balancing the traffic demands on portions of the European airspace while ensuring that their capacity constraints are satisfied.  The traffic demand of a portion of airspace is the hourly number of flights planned to enter it, and its capacity is the upper bound on this number under which air-traffic controllers can work.  Currently, the only form of demand-capacity balancing we allow is ground holding, that is the changing of the take-off times of not yet airborne flights.  Experiments with projected European flight plans of the year 2030 show that already this first form of demand-capacity balancing is feasible without incurring too much total delay and that it can lead to a significantly better demand-capacity balance.
\end{abstract}

\section{Introduction}

The objective of \emph{air traffic management} (ATM) is to ensure a safe, fair, and efficient flow of air traffic, under minimal environmental impact, subject to constraints on aircraft separation, airspace capacity, and airport capacity.  The mission of EuroControl, the \emph{European Organisation for the Safety of Air Navigation} (\href{http://www.eurocontrol.int/}{www.eurocontrol.int}), is to promote the harmonisation of the different national ATM systems.  EuroControl is the counterpart of the \emph{Federal Aviation Administration} (FAA) of the USA.

\subsection{Air Traffic Management in Europe}

Current ATM systems in Europe are fragmented and already stretched to the limit, and hence unable to cope with the traffic volume foreseen for the year 2020 and beyond.  Toward ensuring the required sustainability, an ambitious research effort at pan-European level and under the leadership of EuroControl is now underway to develop a fully integrated next-generation ATM system for the so-called \emph{Single European Sky} (\href{http://www.sesarju.eu/}{www.sesarju.eu}).  The context of this paper is our project on long-term innovative research with the \emph{EuroControl Experimental Centre} (EEC) in Br\'etigny, France.

Today, on the one hand, flight \textbf{planning} is made globally, at the strategic and tactical levels, for all $38$~EuroControl countries by the \emph{Central Flow Management Unit} (CFMU) of EuroControl, upon negotiation with the airlines but without sufficient effort at avoiding traffic bottleneck areas.  On the other hand, flight \textbf{control} takes place more locally, within regional air-traffic control centres (ATCC), but without a super-regional view when flight re-planning has to be done.

The operations of an ATCC rest upon a partition of its civilian airspace into \emph{sectors}, that is three-dimensional (possibly concave) polygonal regions of airspace that are stacked at various altitudes (and that often do not follow national boundaries).  For each sector, a pair of \emph{air-traffic controllers} (ATCo) try to ensure the sector capacity and aircraft separation constraints, at the operational level, and without necessarily physically being at the ATCC headquarters.  For instance, the three BeNeLux countries and small bordering parts of Germany, France, and the North Sea are covered by the Maastricht ATCC in the Netherlands, whose airspace is partitioned into about $5$~or $6$~sectors at every layer, with $2$~or $3$~layers from the ground to upper airspace.

Because of this fragmented mode of operation, the capacity of an ATCC is limited by its sector with the smallest capacity, the \emph{capacity} of a sector being defined as the maximum hourly number of flights that may enter it.  Indeed, the total capacity of an ATCC could be raised by an early identification of the traffic bottleneck areas and a redoing of the flight plans such that the traffic demand is more evenly balanced between its sectors.  This has triggered the investigation of concepts dealing with \emph{multi-sector planning}, where tools are developed to predict the traffic demand over several sectors and to manage the overall demand by anticipating peaks and proposing alternate plans.

Furthermore, a long-term goal is to abandon the existing sectors and reorganise the entire airspace of the \emph{European Civil Aviation Conference} (ECAC) by starting from a three-dimensional grid of same-sized box-shaped \emph{cells} laid over the whole European continent and some nearby countries.  For instance, in our experiments, using a grid of cells whose dimensions are $75$~nm $\times$ $75$~nm $\times$ $125$~FL,\footnote{1 nautical mile = 1.852 km = 1.15 miles; 1 flight level = 30.48 m = 100 feet.} we have $4$~layers with a total of $4600$ ECAC cells, whose capacities are initially uniformly set to $40$~entering flights per hour, to be enforced with a time step of $5$~minutes.

\subsection{Dynamic Demand-Capacity Balancing}

This paper presents the first results on \emph{dynamic demand-capacity balancing} at ECAC level, that is the dynamic modification of ECAC flight plans so as to satisfy the capacity constraints of all cells over a given time interval at minimal cost, thereby avoiding intolerable peaks of demand (and hence ATCo workload), as well as to balance the demands on all cells, thereby avoiding unacceptable dips of demand and unfair discrepancies between demands on the cells.  There are many ways of modifying flight plans and defining the cost of such changes, and we postpone that discussion for a paragraph, so as not to disturb the presentation of the general objective.  We are here only interested in aircraft that follow planned routes, rather than performing free flight.

The tactical rolling-horizon scenario considered is as follows.  At a given moment, suitably called $\now$ below, the (possibly human) demand-capacity manager queries the predicted traffic demands under the given flight plans for the cells of the entire ECAC airspace over a time interval that is some $30$ to $60$ minutes long and starts some $3$ hours after $\now$.  A look-ahead much below $3$ hours would not give enough time for the implementation of some of the necessary re-planning.  A look-ahead much beyond $3$ hours would incur too much uncertainty in trajectory prediction (due to unpredictable weather conditions, say), and hence in demand prediction.  If there are demand peaks, dips, or discrepancies that warrant interference, then the demand-capacity manager launches a re-planning process that suitably modifies the current flight plans.  This process is to be repeated around the clock.  For this to work, the time spent on re-planning should be very short, and the implementation effort of the modified plan should be offset by the resulting demand reductions and re-balancing among the cells.

We now come to the announced discussion of how flight plans can be modified and how the cost of such changes can be defined.  In this paper, we only consider \emph{ground holding}, that is the changing of the take-off time of a not yet airborne flight, with the rest of its plan shifted forward accordingly.  For instance, the EEC suggested that our first experiments should allow ground holding by an integer amount of minutes within the range $[0,\dots,120]$.  Hence the total cost of such flight plan modifications could be the total delay incurred by all ground holding, and this cost is to be minimised.  Our \emph{assumption} here is that take-off times can be controlled with an accuracy of one minute, which is realistic nowadays.  Under this choice, a \emph{flight plan} for a given flight $f$ is reduced to a time-ordered sequence of pairs $(t,c)$, indicating that $f$ enters cell $c$ at time $t$.  Also, a modified flight plan is then obtained by adding a delay $\holding[f]$ to all the cell entry times of the original flight plan of $f$, hence there is a decision variable $\holding[f] \in [0,\dots,120]$ for every flight $f$ of the original set of flight plans.  Other flight plan modifications are discussed in the discussion of future work in Section~\ref{sect:concl}.

In reality, there is another objective function, as the traffic demands on the cells should also be balanced over time and space.  At the time of writing, it is not known yet how to combine any balancing cost with the ground holding cost into a single objective function, so we will just experiment with them separately.  Quantifying the balancing cost may be done in many different ways, and we have decided to start with the standard deviation of the demands on the $4600$ cells.

In short, the \emph{optimisation problem} of the present work can now be stated as follows: Given a set $F$ of flight plans, a set $C$ of airspace cells, and three moments $\now < s < e$ in time, return a modification (only by ground holding) of $F$ of minimal cost, such that the traffic demands on the cells of $C$ never exceed their capacities within the time interval $[s,\dots,e]$.  In practice, an allocated amount $\maxIter$ of iterations is also given, and we want the best such flight plan modifications that can be computed within $\maxIter$ iterations.

\subsection{Contributions and Organisation of this Paper}

This paper makes the following main \emph{contributions}.  To our knowledge, this is the first time that multi-sector planning is attempted on the entire European airspace (and with even many more than the present $20000$ flights per day) and the first time that CBLS is applied to an ATM problem.  We plan to make our constraints and objective functions progressively more realistic: see the discussion of related and future work in Section~\ref{sect:concl}.

The rest of this paper is organised as follows.  In Section~\ref{sect:model}, we summarise a constraint program modelling the dynamic demand-capacity balancing problem in a multi-sector-planning framework.  In Section~\ref{sect:exp}, we report on the experiments we made with that program.  Finally, in Section~\ref{sect:concl}, we conclude, discuss related work, and outline future work.

\section{Model}
\label{sect:model}

Our constraint model is written in \emph{Comet} \cite{Comet:book}, an object-oriented constraint programming language with a back-end for constraint-based local search (available at \href{http://www.dynadec.com/}{www.dynadec.com}).  Some pre-processing (described below) via \emph{MySQL} is made on the raw instance data, which consists of a possibly very large set of flight plans.

\subsection{Parameters and Notation}
\label{sect:parameters}

\begin{figure}[t]
  \begin{center}
    \includegraphics[width=130mm]{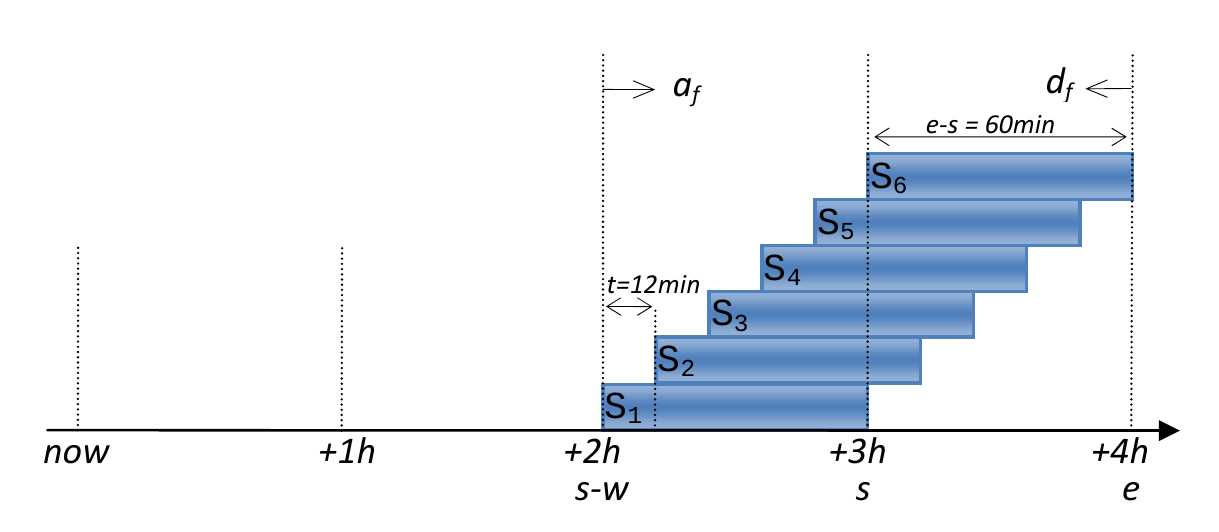}
  \end{center}
  \vspace{-5mm}
  \caption{Sliding windows and the time-line}
  \label{fig:slidingwindows}
\end{figure}

The following notation is used to express the pre-processing and model, all times being expressed as integer amounts of minutes since some common origin:
\begin{align*}
  C &= \text{set of cells of some airspace} \\
  F &= \text{set of flights} \\
  d_f &= \text{planned departure / entry time of flight $f$ in / into the airspace} \\
  a_f &= \text{planned arrival / exit time of flight $f$ in / from the airspace} \\
  C \supseteq C_f &= \text{set of cells entered by flight $f$} \\
  e_{f,c} &= \text{entry time of flight $f$ into cell $c$} \\
  \now &= \text{time of launching the re-planning} \\
  \now < s &= \text{start time of interval for re-planning (typically $3$ hours after $\now$)} \\
  s < e &= \text{end time of interval for re-planning (typically $30$ to $60$ minutes after $s$)} \\
  w &= \text{length of time window over which metric is defined (60 minutes for capacity)} \\
  t &= \text{time step at which metric is checked (a divisor of $e-s$, typically $5$ minutes)} \\
  m &= (e-s)/t = \text{number of time steps along the interval for re-planning} \\
  \maxDelay &= \text{maximum amount of ground holding per flight (currently 120 minutes)} \\
  \capacity &= \text{capacity of a cell ($40$ entering flights per hour, currently for all cells)}
\end{align*}
With the capacity constraint to be enforced every $t$ minutes within the time interval $[s,\dots,e]$ for re-planning, the number of flights entering a given cell has to be measured over a sliding window of $w$ minutes that overlaps with $[s,\dots,e]$.  Sliding window $\Sr$ (where $0 \leq r \leq m$) is the right-open time interval defined by (see Figure~\ref{fig:slidingwindows}):
\begin{equation*}
  \Sr = [s - w + r \cdot t, \dots, s + r \cdot t[
\end{equation*}
that is $S_0 = [s-w,\dots,s[$,~ $S_1=[s-w+t,\dots,s+t[$,~ etc, until $S_m=[e-w,\dots,e[$.

\subsection{Pre-Processing}

The pre-processing derives a few sets from the flight plans in order to express our constraint model and restrict search to only those flights and cells that are affected by the time interval $[s,\dots,e]$ for re-planning:
\begin{itemize}
\item \emph{Relevant Flights} is the set of flights that are planned to depart / enter by the end $e$ of the last sliding window and to arrive / exit upon the beginning $s-w$ of the first sliding window, as otherwise they cannot be airborne in any sliding window of the interval for re-planning (see Figure~\ref{fig:slidingwindows} again):
  \begin{equation*}
    \RF = \{ f \in F \mid d_f \leq e \And a_f \geq s-w \}
  \end{equation*}
Note that the flights that can be ground-held so as to be airborne in some sliding window of the interval for re-planning need not be considered relevant, as they could only worsen the demands on the cells and as the objective of minimising the total amount of ground holding should set their ground holding to zero anyway.
\item \emph{Relevant Airborne Flights} is the set of relevant flights that were planned to depart / enter by $\now$ and can thus not be ground-held:
  \begin{equation*}
    \RAF = \{ f \in \RF \mid d_f \leq \now \}
  \end{equation*}
\item \emph{Relevant Waiting Flights} is the set of relevant flights that are planned to depart / enter after $\now$ and are thus the only ones that can be ground-held:
  \begin{equation*}
    \RWF = \{ f \in \RF \mid d_f > \now \}
  \end{equation*}
\item \emph{Sliding Window Scheduled Flights} is the set of relevant waiting flights planned to contribute demand during a given sliding window $\Sr$ (where $0 \leq r \leq m$):
 \begin{equation*}
    \begin{array}{l}
      \SFr = \{ f \in \RWF \mid c \in C_f \And e_{f,c} \in \Sr \}
    \end{array}
  \end{equation*}
\item \emph{Sliding Window Potential Flights} is the set of relevant waiting flights that can be ground-held in order to contribute demand during a given sliding window $\Sr$ (where $0 \leq r \leq m$):
  \begin{equation*}
    \begin{array}{l}
      \PFr = \{ f \in \RWF \mid c \in C_f \And 
        s - w + r \cdot t - \maxDelay \leq e_{f,c} \leq s - w + r \cdot t \}
    \end{array}
  \end{equation*}
\item \emph{Relevant Cells} is the set of cells containing some possibly ground-held relevant waiting flight in some sliding window:
  \begin{equation*}
    \RC = \{ c \in C_f \mid f \in \RWF \And
    s - w - \maxDelay \leq e_{f,c} \leq e \}
  \end{equation*}
\end{itemize}
Let $P$ be a 2d array such that $P[r,c]$ is initialised during the pre-processing phase to be the \emph{known} part of the demand on relevant cell $c$ during sliding window $\Sr$, that is the number of relevant \emph{airborne} flights that are inside $c$ during $\Sr$.  The \emph{unknown} part of that demand depends on the amount of ground holding imposed on the relevant \emph{waiting} flights.

\subsection{The Decision Variables and the Initial Assignment}

Let $\holding$ be a 1d array of decision variables in the integer domain $[0,\dots,\maxDelay]$, such that $\holding[f]$ is the time delay (duration of ground holding) for the relevant waiting flight $f$.  All delays are set to $0$ in the initial assignment, as we want to ground-hold as few flights as possible.

\subsection{The Constraints}

We currently only have the capacity constraints on all relevant cells during all sliding windows:
\begin{equation} \label{atmost_constraint}
  \begin{array}{c}
    \forall r~ (0 \leq r \leq m) ~.~
    \forall c \in \RC ~.~ \\
    P[r,c] +
    \sum_{f \in \SFr \cup \PFr}
    {\color{red}(} e_{f,c}+\holding[f] \in \Sr {\color{red})} \leq \capacity
  \end{array}
\end{equation}
For a sliding window $\Sr$  and a relevant cell $c$, this constraint takes a flight $f$ from the union of the scheduled and potential relevant flights of $\Sr$ and reifies (for those who view this document in colour: by means of the {\color{red}red} parentheses, which cast the truth of the embraced constraint into $1$ and its falsity into $0$) the interval membership to $\Sr$ of the actual entrance time (under ground holding) of $f$ into $c$.  The sum of the reified memberships for all such flights is the unknown demand on $c$ during $\Sr$; augmented with the known demand $P[r,c]$, the obtained \emph{total demand} must never exceed the capacity (of the cell).

Hence there would be as many as $(m+1) \cdot \card{\RC}$ constraints of the form~(\ref{atmost_constraint}), each involving $\card{\SFr \cup \PFr}$ decision variables and as many reified interval membership constraints for the sliding window $\Sr$.  However, on the ECAC airspace, $\card{\RC}$ easily goes into the thousands under the current values of the problem parameters.  Fortunately, in practice, it is possible to remove about $85\%$ of these constraints without losing any feasible solutions.  Indeed, the constraints involving at most $\capacity$ flights can be removed, since their total demand cannot exceed $\capacity$ (recall that the total demand includes the unknown demand of the relevant waiting flights, which may or may not be ground-held sufficiently so as to enter the considered cell at the considered sliding window).  In other words, the constraint~(\ref{atmost_constraint}) can safely be replaced by:
\begin{equation} \label{atmost_constraint2}
  \begin{array}{c}
    \forall r~ (0 \leq r \leq m) ~.~
    \forall c \in \RC ~.~ \\
    \textbf{if}~ P[r,c] + \card{\SFr \cup \PFr} > \capacity \\
    \textbf{then}~
    P[r,c] +
    \sum_{f \in \SFr \cup \PFr}
    {\color{red}(} e_{f,c}+\holding[f] \in \Sr {\color{red})} \leq \capacity
 \end{array}
\end{equation}
The model can further be shrunk by removing all the relevant cells that are only entered by relevant airborne flights.

Our implementation uses the \emph{atmost} constraint instead of reification, as seen in Algorithm~\ref{algo:constraint}.

\begin{algorithm}[t]
\begin{verbatim}
forall(r in 0..m){
  forall(c in RC){
    currF = collect(f in cellFlight{c}:   // currF = SF_r union PF_r
      f.enterCell >= (now + r*t) && f.enterCell < (s + r*t)
      ) f;
  if (P[r,c] + currF.getSize() > cap)
    S.post(atmost(cap-P[r,c],
      all(f in currF)
      (f.enterCell + deltaT[f]) >= (s - w + r*t) &&
      (f.enterCell + deltaT[f]) <  (s + r*t)
    ));
  }
}
\end{verbatim}
  \caption{Optimised constraint model}
  \label{algo:constraint}
\end{algorithm}

\subsection{The Objective Function}

The capacity constraints are considered soft during the search (in the sense that they may be violated by moves), hence the objective function must take both flight delays and constraint violations into account in order to ensure that the solution has minimum cost.  The objective function to be minimised can be defined with weighted terms as follows:
\begin{equation} \label{objective}
  w \cdot \sum_{f \in \RWF} \holding[f]
  + v \cdot \violations(\text{constraints}~(\ref{atmost_constraint2}))
\end{equation}
where $w$ and $v$ are dynamically changing integer weights, both initially $1$ and maintained by the search procedure (see Section~\ref{sect:search}) to minimise the sum of the delays.  When the objective function cannot be improved, the values of the weights will be increased linearly.  After finding the first solution, the diversification increases the violations and decreases the sum of the delays, so the balance between these weights must be reset to the initial values.

In case we also want to re-balance the demands on the cells, we have to add a weighted term to the objective function~(\ref{objective}), namely
\[
   w' \cdot (\sigma_{c \in \RC,~0 \leq r \leq m} \demand[c,r])
\]
where $w'$ is an integer weight, $\demand[c,r]$ is the number of flights entering cell $c$ during sliding window $\Sr$ under the currently chosen delays, and $\sigma_{i \in T} h(i)$ denotes the standard deviation of the expressions $h(i)$ as $i$ ranges over the set $T$.  In first approximation, we will always have $w \cdot w' = 0$, that is we will minimise only one of these terms of the objective function.

\subsection{The Search Procedure and Heuristics}
\label{sect:search}

We here only give a heuristic and meta-heuristic for minimising the total ground holding (when $w \neq 0 \And w'=0$).  Different (meta-)heuristics may have to be designed for (simultaneous) demand balancing (when $w' \neq 0$).

The search procedure consists of a multi-state heuristic and a meta-heuristic.  An event-based tabu search is used together with an exponential diversification in order to escape local minima during the constraint satisfaction phase, as well as an intensification in order to improve the objective function after the satisfaction phase as the meta heuristic.  We devised a heuristic with three states, and the search procedure may switch states while making moves.  This heuristic is quite involved, but it gives better results than the other approaches we tried, including satisfaction with generic tabu search, variable neighbourhood search, variable-depth neighbourhood search, and weighted optimisation.  Among these approaches, only satisfaction with generic tabu search was faster; however it could not minimise the objective function better than our three-state heuristic.  Hence we now discuss only the latter.

As mentioned in Section~\ref{sect:model}, an optimal solution must satisfy all the constraints and minimise the sum of all delays.  \emph{Furthermore}, an optimal solution is expected to have a maximum of flights with a zero delay, and the percentage of flights with a given delay is expected to decrease as the delay increases.  Indeed, previous studies as well as discussions with CFMU and aircraft operators have shown that it is preferable to have many flights with short delays than a few flights with long delays of the same sum.  In our heuristic, we make use of this feature to narrow down the neighbourhood by first selecting a delay $d$ in the domain $[0,\dots,\maxDelay]$ and then looking for the flight $f$ that achieves the largest violation decrease if $\holding[f]$ is assigned $d$.  The selection process of $d$ cannot be random: it should follow the exponential probability to force the variables to fit best into the expected solution.

We can define the required exponential probability distribution function by dividing the geometric series by the sum of its term:
\begin{equation} \label{expprobability}
  f(x,y) = \dfrac{x^y \cdot (x-1)}{x^{n+1}-x^m},
  \text{~with~} m \leq y \leq n \And \sum_{y=m}^{n} f(x,y) = 1
\end{equation}
where $m$ and $n$ are the powers of the first and last terms in the geometric series, respectively; these values set the \emph{scale} of the probability distribution.  Variable $x$ is a real number called the \emph{ratio} of the series; it affects the speed of growth of the function.  Variable $y$ is an integer ranging from $m$ to $n$.  The delay domain $[0,\dots,\maxDelay]$ being potentially large, we divide it into ten-minute-steps by making $y$ range from $m=1$ to $n=12$ when $\maxDelay=120$; the ratio was experimentally set to $x=1.3$.  Algorithm~\ref{algo:heuristic} depicts the \emph{Comet} implementation of the heuristic.

\begin{algorithm}[t]
\begin{verbatim}
float x = 1.3;
float pr2[i in 1..12] = (x^i)*(x-1.0)/(x^(13.0)-x);
Closure step = closure {
  if(state == 1){
    selectPr(i in 1..12)(pr2[i])
      select(d in ((12-i)*10)..((13-i)*10): d>0)
        selectMin(f in RWF, ad = S.getAssignDelta(deltaT[f],d): 
          tabu[f] <= it && S.violations(deltaT[f]) > 0 &&
          deltaT[f] != d && ad < 0)(ad){
            deltaT[f] := d;
        }
  }
  else if(state == 2){
    selectMax(f in RWF: tabu[f] <= it && S.violations(deltaT[f]) > 0)
      (S.violations(deltaT[f])){
      selectMin(d in 0..g, ad = S.getAssignDelta(deltaT[f],d):
      deltaT[f] != d && ad < 0)(ad,d){
        deltaT[f] := d;
      }
    }
  }
  else if(state == 3){
    selectMin(f in RWF, d in 0..g,
    ad = S.getAssignDelta(deltaT[f],d): tabu[f] <= it &&
    S.violations(deltaT[f]) > 0 && deltaT[f] != d && ad < 0)(ad,d){
      deltaT[f] := d;
    }
  }
};
\end{verbatim}
  \caption{The multi-state heuristic}
  \label{algo:heuristic}
\end{algorithm}

In the first state, when the violation of the constraints falls below an experimentally determined threshold of $300$, the search procedure switches to the second state, which selects the most violated variable and tries to minimise the violation by using the minimum value in its domain.  This state reduces the violation of the constraints very quickly but disregards the total delay, whereas the first state aims at reducing the total delay.  Finally, when the violation of the constraints falls below an experimentally determined threshold of $5$, the third state is activated to expand the neighbourhood to all the variables, so that it is guaranteed that the next move is the best one toward the solution, as long as we are not at local minimum.  The slow performance of the last state will not affect the heuristic since it only applies to the last few violated variables.

The meta-heuristic employs diversification to escape local minima.  The search procedure is depicted in Algorithm~\ref{algo:search}.  The diversification is also moving toward a better objective function by selecting a set of flights and setting their delays to zero.  The selection is not totally random as it uses a reverse sequence of probabilities derived from~(\ref{expprobability}) with a ratio of $1.5$. Once the constraints are satisfied, the diversification level is set to a higher value in order to ensure more modification to the flights' delays, and the state changes according to the new violations.  This process continues to satisfy the constraints again, while the objective function is monitored to store the best solution so far before timing out or maximum number of iteration is reached.

\begin{algorithm}[t]
\begin{verbatim}
while(it<maxIter){
  call(step);
  if (oldViol == S.violations()) steady++;
  else steady = 0;
  if (S.violations() == 0){
    maxDiverse=largeSteps;
    state := 1;
    forall(f in RWF) tabu[f] = 0;
    if (obj < bestSolution){ 
      bestSolution = obj; 
      sol = new Solution(m);
    }
  }
  else {
    maxDiverse = smallSteps;
    if (S.violations() <= 5) state := 3;
    else if (S.violations() <= 300) state := 2;
  }
  if (steady == diversifyLevel){
    with atomic(m){
      forall(i in 0..maxDiverse){
        selectPr(i in 1..12)(pr2[i]){
          select(f in RWF: deltaT[f] > ((i-1)*10) && deltaT[f] <= (i*10)){
            deltaT[f] := 0;
    }}}}
    steady = 0;
  }
  oldViol = S.violations();
}
sol.restore();
\end{verbatim}
  \caption{The search procedure and meta-heuristic}
  \label{algo:search}
\end{algorithm}

It is worth mentioning that, for a problem on such a large number of variables and constraints, achieving a smaller neighbourhood requires much more time per move, which is not acceptable in real-time conditions.  Therefore, a simpler yet faster search procedure can reach better solutions.  In our case, the algorithm can produce acceptable results in a very short time but it will not be able to converge to the optimal solution given enough time.

\section{Experiments}
\label{sect:exp}

The EEC has provided us with a generated data-set of foreseen traffic of the year 2030, namely approximately $50000$ flights per day, including their entry times into the currently considered $4600$ cells and $1200$ airports of the ECAC airspace.

The experiments were performed with \emph{Comet} (revision 2 beta) under Linux Ubuntu 9.0 (32 bit) on an Intel Core 2 Duo T7300 2.0GHz with 2MB cache and 4GB RAM (of which only 2GB are available to \emph{Comet}).  The code is allocating more than 1GB memory to solve the most complex instance.

Our experimental results are based on two time intervals for re-planning.  The first chosen interval is from $s=9$~pm to $e=10$~pm, which is a reasonably busy hour.  The second chosen interval generates the most congested cells and corresponds to the flights departing from $s=5$~pm to $e=6$~pm; there are cells in this interval entered by more than $250$ flights per hour. Table~\ref{tab:stats} shows statistics on the cell demands for these two intervals, taken every $t=12$ minutes in a sliding window of $w=60$ minutes.  These statistics are measured before and after running the minimisation of the total delay, with $\now$ preceding $s$ by $3$ hours, $\maxDelay=120$ minutes of maximum ground holding, $w'=0$ (no demand re-balancing in the objective function), and $\maxIter=40000$ iterations allocated; the minimum and median values are the same before and after optimisation.  Note that all mean values have shrunk, except for the last sliding window of the first interval, because many flights were delayed to fall into that window.  The last two columns confirm that the demands on the cells can indeed hugely exceed their capacity (maximum $\capacity=40$ flight entries per hour), but also show that this capacity can actually be enforced by ground-holding flights for up to $\maxDelay=120$ minutes. 

\begin{table}[t]
  \begin{center}
\scalebox{0.95}{
    \begin{tabular}{|r|r|r|r|r|r|r|r|r|r|r|r|r|r|}
      \hline
      & \multicolumn{2}{|c|}{Mean} & \multicolumn{2}{|c|}{Std Dev} & \multicolumn{2}{|c|}{Variance} & Min & Med & \multicolumn{2}{|c|}{Max} \\
      \hline
      Sliding window & before & after & before & after & before & after & both & both & before & after \\
      \hline
      08:00--09:00~pm &  9.442 & 7.692 & 15.538 & 10.200 & 241.416 & 104.030 & 0 & 3 & 133 & 40 \\
      08:12--09:12~pm &  8.792 & 7.114 & 14.412 &  9.355 & 207.692 &  87.519 & 0 & 3 & 121 & 40 \\
      08:24--09:24~pm &  8.204 & 6.735 & 13.566 &  8.941 & 184.042 &  79.940 & 0 & 3 & 123 & 40 \\
      08:36--09:36~pm &  7.535 & 6.515 & 12.525 &  8.879 & 156.867 &  78.844 & 0 & 3 & 117 & 40 \\
      08:48--09:48~pm &  6.842 & 6.344 & 11.525 &  9.104 & 132.826 &  82.884 & 0 & 3 & 108 & 40 \\
      09:00--10:00~pm &  6.233 & 6.271 & 10.765 &  9.526 & 115.888 &  90.752 & 0 & 2 & 106 & 40 \\
      \hline 
      04:00--05:00~pm & 12.393 & 8.386 & 22.404 & 10.417 & 501.919 &  108.51 & 0 & 4 & 235 & 40 \\ 
      04:12--05:12~pm & 12.144 & 7.994 & 22.042 &  9.609 & 485.863 &  92.332 & 0 & 4 & 235 & 40 \\ 
      04:24--05:24~pm & 11.992 & 7.853 & 22.084 &  9.445 & 487.710 &  89.216 & 0 & 4 & 257 & 40 \\ 
      04:36--05:36~pm & 11.754 & 7.860 & 21.783 &  9.514 & 474.478 &  90.513 & 0 & 4 & 243 & 40 \\ 
      04:48--05:48~pm & 11.513 & 8.129 & 21.580 & 10.128 & 465.707 & 102.570 & 0 & 4 & 245 & 40 \\ 
      05:00--06:00~pm & 11.353 & 8.705 & 21.375 & 11.485 & 456.874 & 131.900 & 0 & 4 & 245 & 40 \\ 
      \hline
    \end{tabular}
}
  \end{center}
  \caption{Statistics on the ECAC cell demands before and after minimisation of the total delay on two chosen time intervals ($s=9$ pm to $e=10$ pm, and $s=5$ pm to $e=6$ pm) that are $3$ hours ahead, with capacity enforced within $[s,\dots,e]$ every $t=12$ minutes over a window of $w=60$ minutes, with a maximum $g=120$ minutes of ground holding, and with $w'=0$ (no demand balancing in the objective function)}
  \label{tab:stats}
\end{table}

\begin{figure}[t]
  \begin{center}
    \includegraphics[width=140mm]{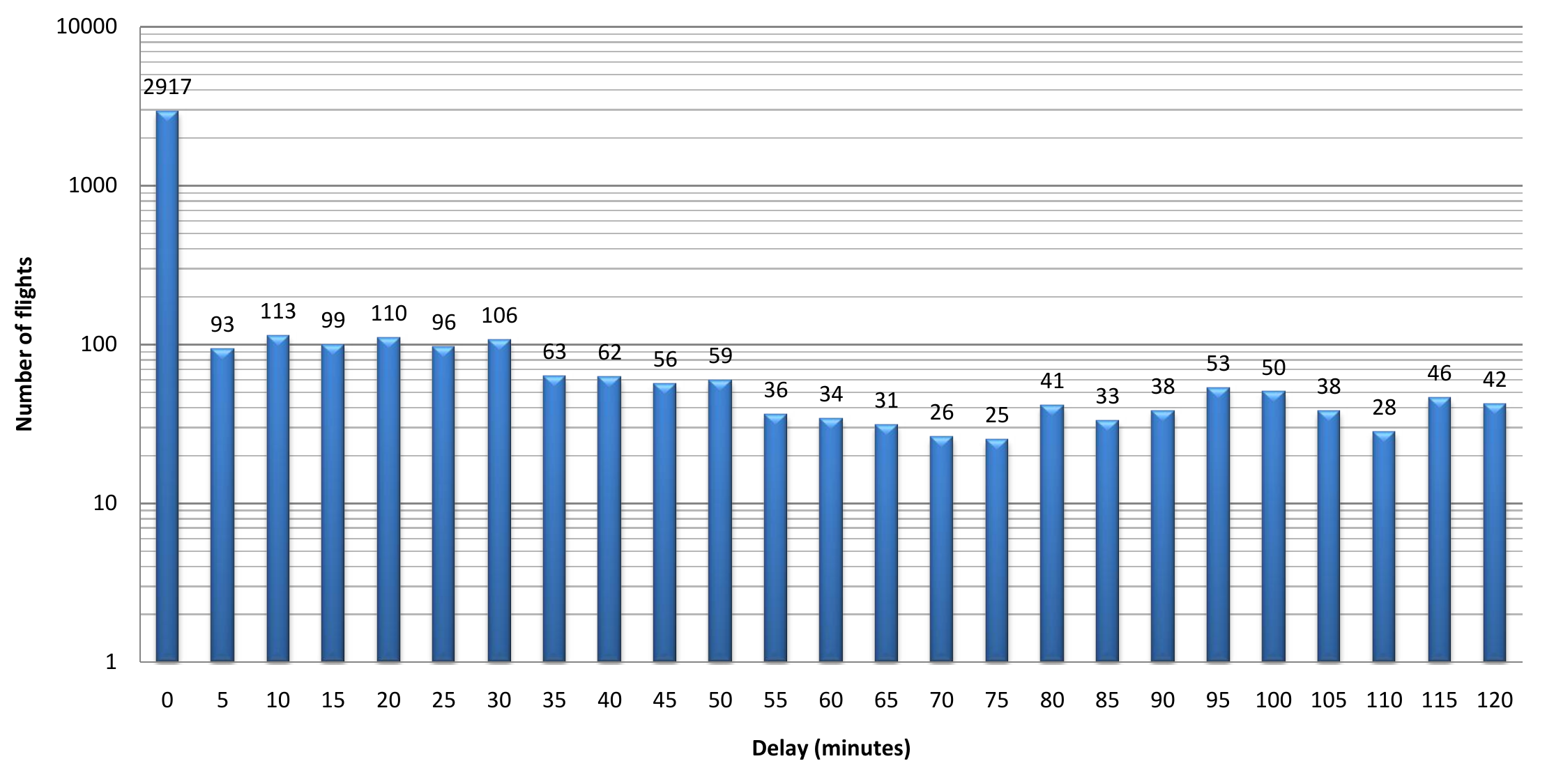}
  \end{center}
  \vspace{-5mm}
  \caption{Numbers of delayed relevant flights for each delay up to $\maxDelay=120$ minutes, bundled into $5$-minute-intervals}
  \label{fig:histogram}
\end{figure}

The distribution of the delays among the relevant flights for the first interval is presented in Figure~\ref{fig:histogram} as a histogram.  Note the logarithmic scale of the vertical axis: the vast majority of the flights are not ground-held at all and, as wanted, the numbers of flights are more or less decreasing when the delays are increasing.

Finally, Table~\ref{tab:results} gives for both intervals the run-time, the actual number of iterations, the numbers of relevant waiting flights (that is the number of decision variables) and relevant airborne flights (which cannot be ground-held) at moment $\now$, the total and average amounts of ground holding for all relevant flights, the change in standard deviation of the demands over all cells, and the initial number of constraint violations.  It is very important to observe that, although the aim was only to minimise the total delay, there also is a $35\%$ (respectively $43\%$) improvement in the standard deviation of the cell demands.  About $70\%$ (respectively $50\%$) of the relevant \emph{waiting} flights need not be ground-held at all; note that this percentage does not include the relevant airborne flights, nor the many flights that could possibly be ground-held so as to be airborne within the interval for re-planning.

\begin{table}[t]
  \begin{center}
\scalebox{0.92}{
    \begin{tabular}{|l|r|r|r|r|r|r|r|r|}
      \hline
      Interval & Run-time & Iterations & $\card{\RWF}$ & $\card{\RAF}$ & Total Delay & Avg Delay & Demand Dev & Violations \\
      \hline
      9--10~pm & 140 sec &  8,000 & 4,295 & 463 &  65,457 min & 13.76 min & $-35\%$ &  9,013 \\
      \hline 
      5--06~pm & 620 sec & 22,000 & 8,806 & 811 & 246,267 min & 25.61 min & $-43\%$ & 45,985 \\ 
      \hline
    \end{tabular}
}
  \end{center}
  \caption{Statistics for the two chosen time intervals: run-times, numbers of iterations, numbers of relevant waiting and airborne flights, total and average delays of all relevant flights, changes in the standard deviations of the cell demands, and initial numbers of constraint violations}
  \label{tab:results}
\end{table}

Among the parameters of our model (see Section~\ref{sect:parameters}),
the maximum amount of ground holding per flight (currently set to
$\maxDelay=120$ minutes) is the most interesting one to play with.
Our first experiments show that it can only be lowered to about
$\maxDelay=100$ minutes without losing the satisfiability of the
capacity constraints, though without significant deterioration of the
run-time or total delay.

Note that the trivial way to meet all the capacity constraints (namely
by delaying a maximum of flights out of the time interval for
re-planning, and thus exacerbating the situation for the next such
interval) would be sub-optimal since we aim at minimising the total
delay.

\section{Conclusion}
\label{sect:concl}

\paragraph{Summary.}
Constraint-based local search (CBLS) offers a very effective medium for \emph{modelling} and efficiently \emph{solving} the problem of optimally balancing the traffic demands on an airspace of adjacent cells, while ensuring that their capacity constraints are never violated.  The demand on a cell is here defined as the hourly number of flights entering it.  The currently only allowed form of re-planning is the changing of the take-off times of not yet airborne flights.  Experiments with projected European flight plans of the year 2030 show that this can already lead to significant demand reductions and re-balancing at little cost.

Ultimately, we hope to show that EuroControl can gain from a tight integration of planning and control.  As discussed in the introduction, currently flight planning is made globally by its CFMU unit a long time in advance but without achieving optimal flow and under data estimates that are almost certainly incorrect, whereas flight control is made locally by its ATCC units when more precise data is available but without a global view, so that na\"ive re-planning takes place.

A manifestation of this lack of integration of planning and control is that the delays reported in Table~\ref{tab:results} and Figure~\ref{fig:histogram} would seem quite unacceptable (for instance, about $20\%$ of the flights incur at least $30$ minutes of delay).  This is due to the poor cell demand distribution of the input flight plans (see Table~\ref{tab:stats}) and is thus an argument for the proposed tighter integration of planning and control.

\paragraph{Related Work.} 
Our previous project~\cite{ASTRA:JATM07} with EuroControl headquarters in Brussels (Belgium) is here extended from five airspace sectors (covering a bit more than the three BeNeLux countries) to the entire ECAC airspace, divided into box-shaped cells rather than the current sectorisation, and (initially) using capacity as a metric rather than some form of air-traffic complexity.  Since this is a very large jump in the size of the input data, and since the technique to be developed even has to cope in real time with traffic volumes of the year 2030 (about $50000$ flights per day, as opposed to the $20000$ flights nowadays), our first idea was to use CBLS instead of the traditional constraint programming (by heuristic-guided global tree search with propagation at every node explored) that we deployed in the previous project.  Since the chosen CBLS modelling language, namely \emph{Comet}, also has constraint programming (by global search) and integer programming back-ends, we plan to keep the model and evaluate the actual necessity of this choice.

The works closest to ours were tested on actual flight data of the year 1995 for all of France~\cite{Junker:ISA} and the year 1999 for one French ATCC~\cite{Barnier:ATM01}.  In further contrast to our present work (which was tested on projected flight data of the year 2030 for the entire ECAC airspace, divided into uniform cells rather than sectors), traditional constraint programming (rather than CBLS) was used toward satisfying the capacity constraints of all sectors, and the objective of the latter work was to minimise the maximum of all the imposed ground holdings (rather than their sum).

Another important related work concerns the airspace of the USA \cite{Sherali:03}.  The main differences with our work are as follows.  They have (at least initially) static lists of alternative routes to pick from for each flight and do not consider changing the time plans, whereas we currently only dynamically modify time plans.  Their sector workload constraints limit the average number of flights in a sector over a given time interval (like our demand) but also the number of PIPs.  No multi-sector planning is performed to reduce and re-balance the workloads of contiguous sectors.  However, a notion of airline equity is introduced toward a collaborative decision-making process between the FAA and the airlines.

Finally, there is related work on minimising costs when holding flights (on the ground or in the air), if not re-routing them, in the face of dynamically changing weather conditions \cite{Bertsimas:00}.  The main differences with our work are as follows.  Their objective is cost reduction for airlines and airports, whereas our work is airspace oriented.  They consider ground holding and air holding, whereas we currently only consider ground holding.  Their dynamic re-routing is on the projected two-dimensional plane, whereas we do not re-route yet.  Their sector workload constraints limit the number of flights in a sector at any given time (this is called sector \emph{load} below), but there is no multi-sector planning.

A lot of related work is about dealing with potentially interacting pairs (PIPs) of flights (see~\cite{Barnier:ATM09,Hildum:ICAPS07} for instance).  However, this is an operational issue, whereas our work is at the tactical level, so we do not have to worry about the number of PIPs in the modified flight plans and hence we do not enforce any aircraft separation constraints.

As this brief overview of related work shows, the whole problem of optimal airspace and airport usage by the airlines is very complex, and only facets thereof are being explored in each project.  Our work intends to reveal some new facets, such as multi-sector planning.

\paragraph{Future work.}
At present, this work is only a feasibility study, and this induced some simplifications in our experiments.  For instance, the cells in our grid should have different capacities (rather than the uniform $40$ we experimented with), just like the current sectors, which have roughly the same traffic volumes.  Indeed, demand balancing under equal capacities does not make much sense, witness for instance low-level cells in areas without airports.  However, our model can now be used to fine-tune the individual cell capacities.

We must add side constraints to the current trunk of only capacity constraints in order to make things more realistic.  For instance, since the re-planning happens for a time interval in the future, constraints will be needed to make sure no unacceptable traffic demand is generated \emph{before} that interval.  Additional constraints are also needed to make sure that the flight plan modifications can be implemented sufficiently quickly and cheaply, and that doing so is still offset by the resulting demand reductions and re-balancing among cells.  For instance, the number of flights affected by the changes may have to be kept under a given threshold.  This is where airlines and pilots will probably have to be factored in.

Our constraint model does not enforce any notion of first-planned-first-served, which is a goal for CFMU and the FAA, since the delays would then become very huge.  We will investigate situations where such a fairness notion can be applied nevertheless.

Capacity (the maximum number of hourly flights that may enter a portion of airspace) is not the only metric we can enforce (under a given time step).  Another common ATM metric is \emph{load}, namely the maximum number of flights that can be simultaneously present in the portion of airspace.  In a previous study~\cite{Mercier:load}, it was concluded that load constraints lead to less total delay than capacity constraints.  Ultimately, a \emph{workload} metric should be used (as in our~\cite{ASTRA:JATM07}), estimating the traffic complexity as it is perceived by air-traffic controllers.

Many other forms of flight re-planning can be imagined beyond the ground holding we here experimented with, such as the vertical re-routing of flights along the planned 2D route, the horizontal 2D re-routing along alternative routes (from a list of fixed or dynamically calculated routes), accelerations and decelerations, and air holding.  The cost function has to be adapted accordingly.  Such additional forms of re-planning should only be introduced if they are warranted by additional cost gains and by computational feasibility.

There is a lot of other future work to do before an early prototype like ours can be deployed in a tactical context.  Its main current objective is therefore research strategic, namely to provide a platform where new metrics can readily be experimented with, and where constraints and objective functions can readily be changed or added.  This motivated the choice of constraint programming as implementation technology, since the maintenance of constraint programs is simplified compared to ad hoc programs.

\subsubsection*{Acknowledgements}

This work has been co-financed by the European Organisation for the Safety or Air Navigation (EuroControl) under its Care INO III programme (grant 08-121447-C).  The content of the paper does not necessarily reflect the official position of EuroControl on the matter.  Many thanks to Franck Ballerini, Marc Bisiaux, Marc Dalichampt, Hamid Kadour, and Serge Manchon at the EuroControl Experimental Centre for the definition of the problem, the data-set used, and the feedback on our progress.  Special thanks to the reviewers for pointing out many ways to improve the presentation of this paper.

\bibliographystyle{eptcs}

\end{document}